\documentclass[twocolumn]{rps-esrel2025}
\usepackage{tikz}
\usetikzlibrary{shapes,arrows}
\tikzstyle{decision} = [diamond, draw, fill=blue!20, 
    text width=4.5em, text badly centered, node distance=3cm, inner sep=0pt]
\tikzstyle{block} = [rectangle, draw, fill=blue!20, 
    text width=5em, text centered, rounded corners, minimum height=4em]
\tikzstyle{line} = [draw, -latex']
\tikzstyle{cloud} = [draw, ellipse,fill=red!20, node distance=3cm,
    minimum height=2em]
\usepackage{siunitx}
\def\papername{\jobname}

\usepackage{natbib}

\begin{document}

\markboth{Georgios Katranis et al.}{Dynamic Risk Assessment for Human-Robot Collaboration Using a Heuristics-based Approach}

%%%%%%%%%%%%%%%%%%%%%%%%% Plase keep this command for single column for abstract section.
\twocolumn[
%%%%%%%%%%%%%%%%%%%%%%%%%

\title{Dynamic Risk Assessment for Human-Robot Collaboration Using a Heuristics-based Approach}

\author{Georgios Katranis\textsuperscript{1}, Frederik Plahl\textsuperscript{1,2}, Joachim Grimstadt\textsuperscript{1}, Ilshat Mamaev\textsuperscript{2}, Silvia Vock\textsuperscript{3}, Andrey Morozov\textsuperscript{1}}
\address{\textsuperscript{1} University of Stuttgart, Institute of Industrial Automation and Software Engineering (IAS), Germany. \email{\{first.last\}@ias.uni-stuttgart.de}}
\address{\textsuperscript{2} Proximity Robotics \& Automation GmbH, Pfinztal, Germany. \email{\{last\}@proximityrobotics.com}}
\address{\textsuperscript{3} Federal Institute for Occupational Safety and Health (Bundesanstalt für Arbeitsschutz und Arbeitsmedizin, BAuA), Germany. \email{\{last.first\}@baua.bund.de}}

\begin{abstract} 

Human-robot collaboration (HRC) introduces significant safety challenges, particularly in protecting human operators working alongside collaborative robots (cobots). While current ISO standards emphasize risk assessment and hazard identification, these procedures are
often insufficient for addressing the complexity of HRC environments, which involve numerous design factors and dynamic interactions. This publication presents a method for objective hazard analysis to support Dynamic Risk Assessment, extending beyond reliance on expert knowledge. The approach monitors scene parameters, such as the distance between human body parts and the cobot, as well as the cobot's Cartesian velocity. Additionally, an anthropocentric parameter focusing on the orientation of the human head within the collaborative workspace is introduced. These parameters are transformed into hazard indicators using non-linear heuristic functions. The hazard indicators are then aggregated to estimate the total hazard level of a given scenario. The proposed method is evaluated using an industrial dataset that depicts various interactions between a human operator and a cobot.

\end{abstract}


%%%%%%%%%%%%%%%%%%%%%%%%% Please keep this closing bracket to complete the single column format for abstract.
]
%%%%%%%%%%%%%%%%%%%%%%%%%

\section{Introduction}\label{sec:introduction} 
Human-Robot Collaboration (HRC) is increasingly being deployed in industrial contexts. This trend is underscored by the projected growth of the global collaborative robot (cobot) market, with its size expected to increase exponentially by $2030$ \citep{next_move_strategy_2022}. In HRC, cobots and human operators share a workspace, combining the strengths of robots—such as the ability to perform repetitive tasks with high precision and efficiency—with the advanced cognitive and sensory capabilities of humans \citep{ISO15066}. The emergence of HRC introduces not only opportunities but also a range of challenges. The most critical challenge is ensuring the safety of human operators. To address this issue, it is necessary to conduct a thorough risk assessment prior to the implementation of a robot system. This assessment should evaluate the robot itself, as well as the work environment \citep{ISO10218}. Additionally, collaborative systems need to employ sophisticated methodologies, such as adaptive speed limitations, area monitoring, power and force limitation to achieve safety requirements set forth in the standards \citep{ISO15066}.

Nevertheless, a lot of challenges regarding safety remain unaddressed. 
\cite{Huck_Risk_assesment_survey} argue that risk assessment as recommended by current standards, relying on experience and expert knowledge, is challenging to apply to HRC. The difficulty arises due to the fields inherent complexity, the limited experience in this domain, challenges in predicting human behavior, and the intricacies of estimating collision criticality. 
Among these, lack of experience is often cited as the most significant barrier. Practitioners frequently report insufficient knowledge of various aspects of the domain \citep{aaltonen2019experiences}. Moreover, questions arise regarding which operational safety functions and methods should be applied at different interaction levels in HRC, and which robot parameters should be measured or controlled during each level of interaction \citep{bdiwi2017new}. Finally, \cite{zacharaki2020safety} address the need for mathematical modeling and systematization of existing State-of-the-Art methods in order to allow hazard avoidance prior to accidents. They contend that the establishment of unified metrics and datasets is paramount to enhance safety.

Addressing the identified shortcomings requires a Dynamic Risk Assessment (DRA) method that extends beyond sole reliance on expert knowledge. This paper proposes an objective approach to hazard identification, wherein the hazard level in a given a scenario is dynamically quantified through numerical values. Hazard analysis—a key aspect of risk assessment—is conducted by defining hazard indicators that are based on mathematical functions. This ensures a consistent and objective evaluation of a scenario’s danger level and severity. By providing a structured, data-driven understanding of hazards, the method aims to simplify subsequent risk estimation for system integrators or robotic systems.

The main contributions are twofold. Firstly, an anthropocentric parameter based on human head orientation is introduced. Secondly, based on scene parameters and the anthropocentric parameter, non-linear, heuristic hazard indicators are developed. These indicators are then aggregated into a comprehensive framework for quantifying and assessing hazards, enabling a more consistent and systematic scenario analysis that incorporates human behavior.

\section{Related Work} \label{sec: related_work}

ISO/TR $14121\mathrm{-}2$ provides practical methods for implementing an ISO $12100$-compliant risk assessment process, defining risk indices based on the probability of hazardous events and the severity of potential accidents. While the standard outlines quantitative and semi-quantitative approaches, its methods for calculating risk indices include subjective elements. Initial system assumptions are evaluated using risk graphs, a risk matrix, or numerical scoring, assigning values to harm severity and occurrence probability. Numerical scoring uses numbers, whereas other methods rely on descriptive categories like "very likely" or "rare" to characterize risk factors. The assigned values are then combined to produce a total risk score \citep{ISO14121}. \cite{Kulic_real_time_safety_for_HRI} formulated a numerical danger index to evaluate human-robot interaction risks. The index incorporates parameters such as the distance between robot and human, relative velocity, and effective robot inertia at impact points. These parameters collectively capture the likelihood and severity of a potential collision, computing risk for the closest point on each robot link to the human. \cite{sanderud2015proactive} proposed the creation of a dynamic risk map of the workspace, which incorporates both the likelihood of human presence and the severity of potential consequences. Their model dynamically adjusts based on limb velocity, along with other parameters such as proximity and body part criticality for evaluating risks. Similarly, \cite{dittrich2016robot} developed a Bayesian Network (BN)-based model for dynamic risk estimation. Their model combines local body-part-specific information with global situational factors, such as human attention or posture, using Fuzzy Logic (FL). Another study by \cite{beltran2018fuzzy} combined FL, Artificial Neural Networks (ANN), and Adaptive Neuro-Fuzzy Inference Systems (ANFIS) to compute a Hazard Rating Number. Their model evaluates factors like proximity, speeds, head orientation, and upper body posture. Finally, \cite{Huck_Testing_System_Safety} presents a simplified heuristic metric to identify hazard. In their publication, a deterministic function to compute the risk is defined based on the proximity between human and robot in the workspace.

\section{Heuristics-based Hazard Analysis} \label{sec:heuristics_based_ra}

\subsection{Quantitative Definition of Hazards and Risk}

To formally quantify risks and hazards this work adopts the definition proposed by \cite{kaplan1981quantitative} who define risk as a set of triplets representing all possible risky scenarios:
\begin{equation}
    R = \{ \langle s_\mathrm{i}, p_\mathrm{i}, x_\mathrm{i} \rangle \}
\end{equation}
where $s_\mathrm{i}$ is a scenario identification or description, $p_\mathrm{i}$ is the probability of that scenario, and $x_
\mathrm{i}$ is the consequence or evaluation measure of
that scenario. Naturally, a formal definition of hazard can be defined as:
\begin{equation}
H = \{ \langle s_\mathrm{i}, p_\mathrm{i} \rangle \}
\end{equation}

This definition serves as the basis for the adopted methodology, with two reinterpretations. First, a scenario is treated as a collection of measurable parameters that assess the associated danger. For example, the velocity of a robot can either pose a hazard or remain harmless. In this context, probabilities are not computed for the existence of a dangerous scenario. Instead, a continuous numerical characterization of the danger is undertaken based on the identified parameters, resulting in hazard indicator values. Existing standards aim to create inherently safe systems by defining acceptable parameter ranges for robots to operate within human-compatible limits. Leveraging this fact allows for precise hazard identification within specific velocity ranges. This principle extends to other parameters, such as distance, enabling direct hazard identification without relying on probabilistic estimates. Additionally, incorporating human behavior introduces variability, as actions within the workspace can transform an otherwise safe scenario into a hazardous one.

\subsection{Heuristic Metrics}

To analyze hazards effectively, relevant system parameters must first be identified and made comparable. The proposed method for quantifying a scenario as hazardous is based on heuristic functions. It is inspired by the works of \cite{Huck_Testing_System_Safety} and \cite{Kulic_real_time_safety_for_HRI}. The identified parameters are then mapped from their domain to a range of $[0,1]$ through the heuristic functions. These are non-linear, introducing a skew that emphasizes high-danger scenarios by increasing sensitivity near maximum thresholds and reducing it near minimum values. This framework offers several advantages. It provides a transparent, deterministic method for hazard identification, ensuring all parameters contribute clearly and are adjustable. Unlike machine learning-based models, this approach does not require large annotated datasets, making it ideal for scenarios where such data is unavailable or impractical to obtain. Furthermore, it is highly flexible, allowing for quick adaptation to specific tasks or environmental changes. 

To determine the required functions, the parameters to be monitored must first be established. \cite{ISO15066} identifies essential parameters, including separation distance between human and robot, and robot velocity. Additionally, a new anthropocentric parameter, the Position of Human Head (PHH), is introduced to account for human awareness. The standard emphasizes, that the head and face are particularly vulnerable areas of the human body. Given this sensitivity, it is critical to closely track the position of the human head during operations. Head movement can provide significant insights into potentially hazardous scenarios that would otherwise be safe if a person was attentive.
\subsection{Definition of Hazard Indicators}

\subsubsection{Distance-based Hazard Indicator}

To identify the hazard level of a scenario based on the proximity between a robot and a human, the indicator must reflect that the indicator of harm increases as the distance decreases. Conversely, greater distances reduce the associated indicator. \cite{Huck_Testing_System_Safety} model this relationship using a heuristic negative exponential function, which serves as the foundation for the proposed approach.

The baseline function is augmented with parameters to account for robot-specific constraints and established safety standards. The function is expressed as:
\begin{equation}
    r_\mathrm{D}(d_\mathrm{H}) =
    \begin{cases}
    e^{-\alpha \cdot (d_\mathrm{H} - d_\mathrm{min})}, & \text{if } d_\mathrm{min} \leq d_\mathrm{H} < d_\mathrm{reach}, \\
    0, & \text{if } d_\mathrm{H} \geq d_\mathrm{reach}, \\
    1, & \text{if } d_\mathrm{H} \leq d_\mathrm{min}
    \end{cases}
    \label{eq:distance_risk}
\end{equation}

Parameter $\alpha$ controls the steepness of the curve, $d_\mathrm{H}$ is the distance between the human and the robot, and $d_\mathrm{min}$ denotes the minimum safety distance as specified in ISO/TS 15066:2016.
An additional constraint includes the robot’s maximum reach, $d_\mathrm{reach}$, which defines the physical limit of its workspace.

The parameter $d_\mathrm{min}$ ensures that $r_\mathrm{D}(d_\mathrm{min}) = 1$ when the human is within the minimum safe distance from the robot. It can be determined by using the robot’s stopping time, $T_\mathrm{stop}$, for a given velocity through the inequality: \citep{ISO15066}
\begin{equation}
    \label{eq:inequality_min_safe_dist}
    d_\mathrm{min} \geq v \cdot T_\mathrm{stop}
\end{equation}
$d_\mathrm{reach}$ ensures that $r_\mathrm{D}(d_\mathrm{H})=0$ for distances beyond the robot’s physical reach, as the robot cannot pose any danger to the human outside this range.

\begin{figure}[!htpb]
    \epsfxsize=.001\hsize
    \centerline{\epsfbox{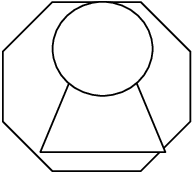}}
    \includegraphics[width=\linewidth]{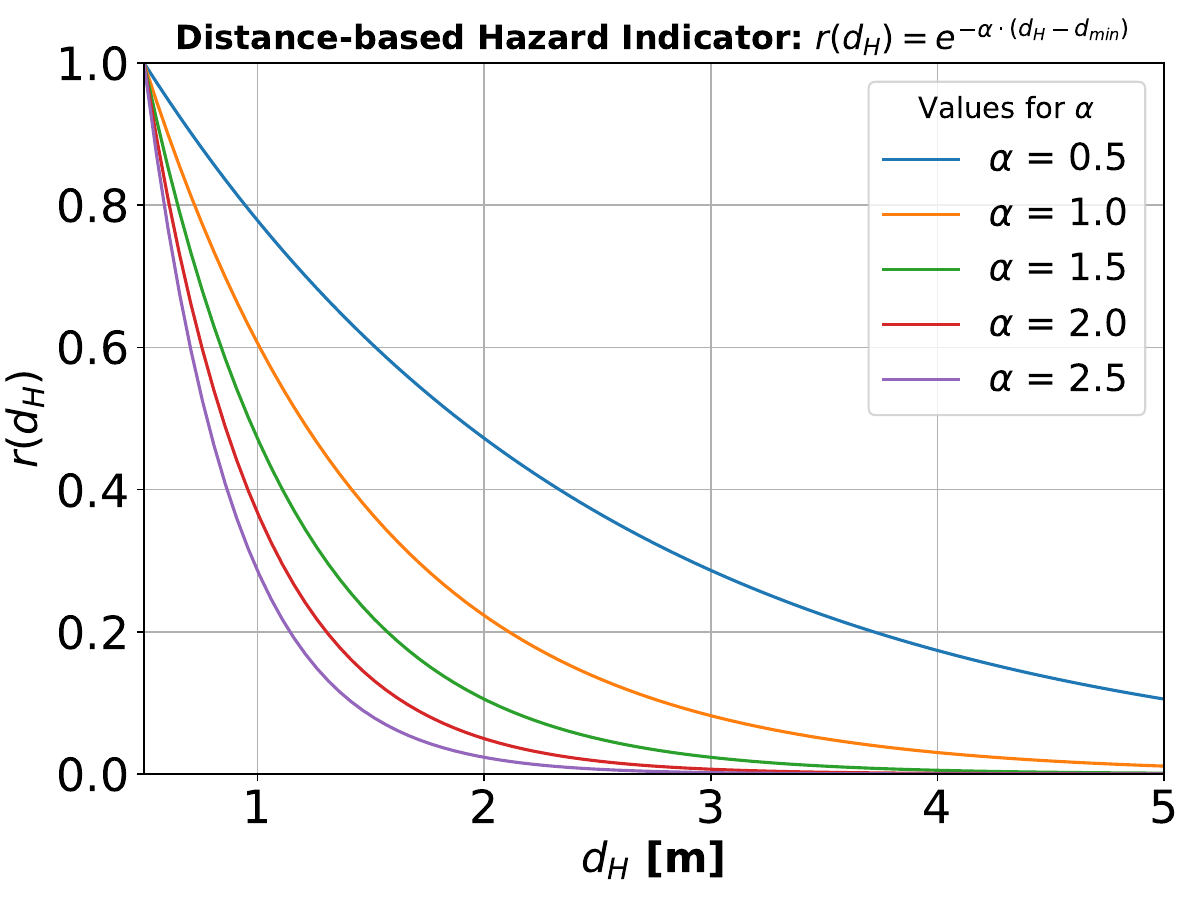}
    \caption{Distance-based hazard indicator for different values of $\alpha$ with $d_\mathrm{min} = 0$.}
    \label{fig:Distance_Metric}
\end{figure}

Fig. \ref{fig:Distance_Metric} presents the plot of the function for different values of $\alpha$. It becomes evident that increasing $\alpha$ makes the function sensitive to small changes in distance. This means that even minor changes lead to a significant increase in the hazard. To determine an appropriate value for $\alpha$, Eq. (\ref{eq:distance_risk}) is solved using the specified constraints. It is evident that the evaluation of distance-based hazards relies entirely on the robot's characteristics and operational parameters.

\subsubsection{Velocity-based Hazard Indicator} \label{sec:vel_based_hazard}

To assess the hazard level based on the robot’s velocity, both its magnitude and direction must be considered. The metric should reflect that an increase in velocity correlates with a higher associated danger, while a decrease in velocity reduces the hazard. \cite{Kulic_real_time_safety_for_HRI} propose a quadratic function to compute a hazard indicator based on the magnitude of velocity, expressed as: 
\begin{equation}
    \begin{aligned}
        r_\mathrm{V}(v) &= k_\mathrm{V} \cdot (v - V_\mathrm{min})^2 \\
           \text{with} \quad k_\mathrm{V} &= \frac{1}{(V_\mathrm{max} - V_\mathrm{min})^2}
    \end{aligned}
    \label{eq:risk_only_magnitude_of_vel}
\end{equation}
$V_\mathrm{min}$ is the lowest threshold velocity below which the associated risk is negligible. It is typically set to $250$\si{\milli\meter\per{\second}} \citep{ISO10218}. $V_\mathrm{max}$ is chosen in accordance with ISO/TS $15066\mathrm{:}2016$.

However, this equation has a limitation. It only partially considers the vector nature of velocity. The directional component is simplified as positive or negative \citep{Kulic_real_time_safety_for_HRI}. To assess danger more accurately, the full directional information must be accounted for, as velocity is not always strictly forward, or backward. \cite{marvel2017implementing} highlight that current standards do not factor in the direction of travel. From a safety perspective, it is assumed that the operator’s movements may follow a worst-case scenario, where the operator moves in the direction of the robot at any moment. To address this, a worst-case-direction vector, $\vec{d}$, is defined representing the direction in which the hazard is expected to increase the most. This is the case when the robot moves directly toward a human operator. The directional component of the robot’s velocity vector $\vec{v}$ in the direction of $\vec{d}$ is computed by projecting the vectors onto each other. Using the relationship given by the dot product and reordering yields:
\begin{equation}
    \cos(\theta) = \frac{\vec{v} \cdot \vec{d}} {\|\vec{v}\| \cdot \|\vec{d}\|}
    \label{eq:projection}
\end{equation}
The term represents the directional influence, indicating how much of $\vec{v}$ is aligned with $\vec{d}$. When $cos(\theta) \approx 1$ , the robot's velocity is highly aligned with the worst-case direction, increasing the hazard. Conversely, when 
$cos(\theta) \approx 0$ or negative, the robot's velocity is less aligned or even moving away from the worst-case direction, resulting in a lower hazard indicator. To normalize this directional influence and ensure the risk metric remains within a bounded range of $[0, 1]$, the following equation is used:
\begin{equation}
    \label{eq:velocity_risk_directional}
    r_\mathrm{V_{directional}}(\theta) = \frac{1+ cos(\theta)}{2}
\end{equation}
The final velocity-based hazard indicator combines the magnitude of the velocity and its directional influence into a weighted sum.
\begin{equation}
        r_\mathrm{V}(v, \theta) = \beta \cdot  r_\mathrm{V}(v) +  (1-\beta) \cdot r_\mathrm{V_{directional}}(\theta)
    \label{eq:velocity_risk}
\end{equation}    
$\beta$ is the weighting factor the range of $[0, 1]$. The initial estimate for $\beta = 3/4$ emphasizing the velocity magnitude. The velocity hazard indicator remains zero, if $v < V\mathrm{min}$.

\begin{figure}[!htpb]
    \epsfxsize=.001\hsize
    \centerline{\epsfbox{ch01f01.eps}}
    \includegraphics[width=\linewidth]{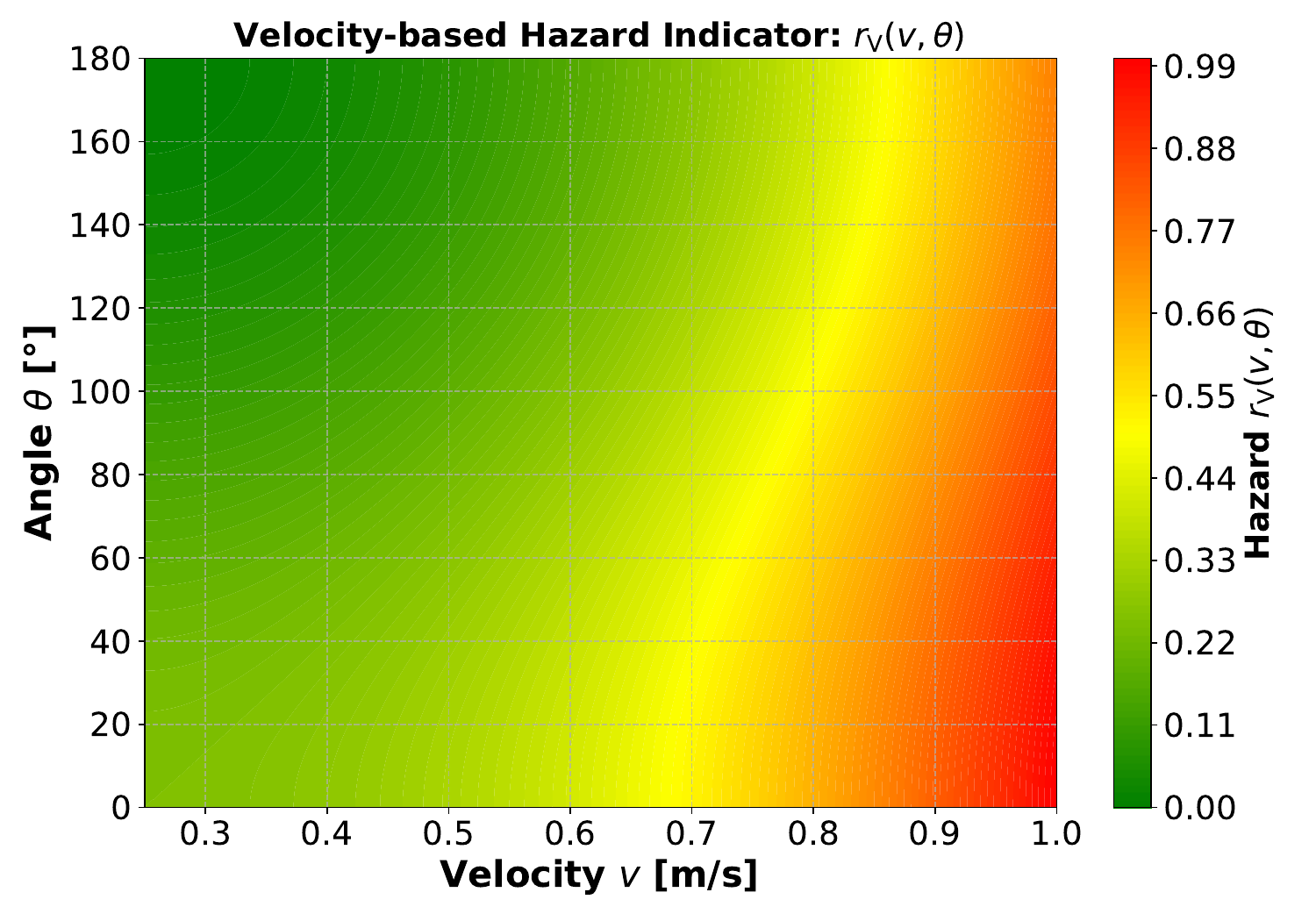}
    \caption{Heat map of the velocity-based hazard indicator.}
    \label{fig:velocity_risk_total}
\end{figure}
Figure \ref{fig:velocity_risk_total} illustrates the hazard indicator as a heat map, plotted based on velocity magnitude and the directional component, represented by the angle $\theta$. The maximum velocity in this plot is set to $V_\mathrm{max}=1$ \si{\meter\per{\second}}. As expected, the hazard is highest when the velocity reaches its maximum value and $\theta = 0\si{\degree}$, as indicated by the red region on the heat map. Conversely, the hazard is lowest when the velocity is minimal and $\theta = 180\si{\degree}$, represented by the green region. Interestingly, the region where the velocity is maximal but indicating the robot is moving away from the human still shows high danger. This is due to the inherent dangers of high-velocity movement, such as the potential for unforeseen circumstances. For example, a robot moving rapidly might accidentally drop an object held by its gripper.

\subsubsection{Position of Human Head-based Hazard Indicator}

The PHH indicator is unique among the proposed hazard indicators as it is entirely anthropocentric and not derived from established standards. It quantifies the deviation of the human head's orientation from an expected reference. Similarly to the velocity-based metric, it is designed to increase with larger deviations in orientation and decrease with smaller deviations. PHH is an angular value and is described through a sigmoid growth curve:
\begin{equation}
    \label{eq:sigmoid_hfov}
    r_\mathrm{PHH}(PHH) = 1 - e^{-(\frac{PHH}{c})^2}
\end{equation}  
$c$ controls the steepness of the curve, enabling fine-tuning to match the desired hazard scaling. To compute the indicator, a reference vector, $\vec{p_\mathrm{ref}}$, is defined, extending from a point on the human head to the robot’s end-effector. This vector represents the ideal direction for human focus during interaction. A current orientation vector, $\vec{p_\mathrm{orient}}$, denoting the actual direction of the human’s gaze, is also constructed. The normalized current orientation vector is projected onto the normalized reference vector. The relationship between the vectors is identical to the one described in Eq. (\ref{eq:projection}). PHH is compured only for the yaw angle, i.e rotation around the z-axis, as it most critically affects the human’s field of view. However, the other two rotational axes can also be monitored separately using the same function.
\begin{figure}[!htpb]
    \epsfxsize=.001\hsize
    \centerline{\epsfbox{ch01f01.eps}}
    \includegraphics[width=\linewidth]{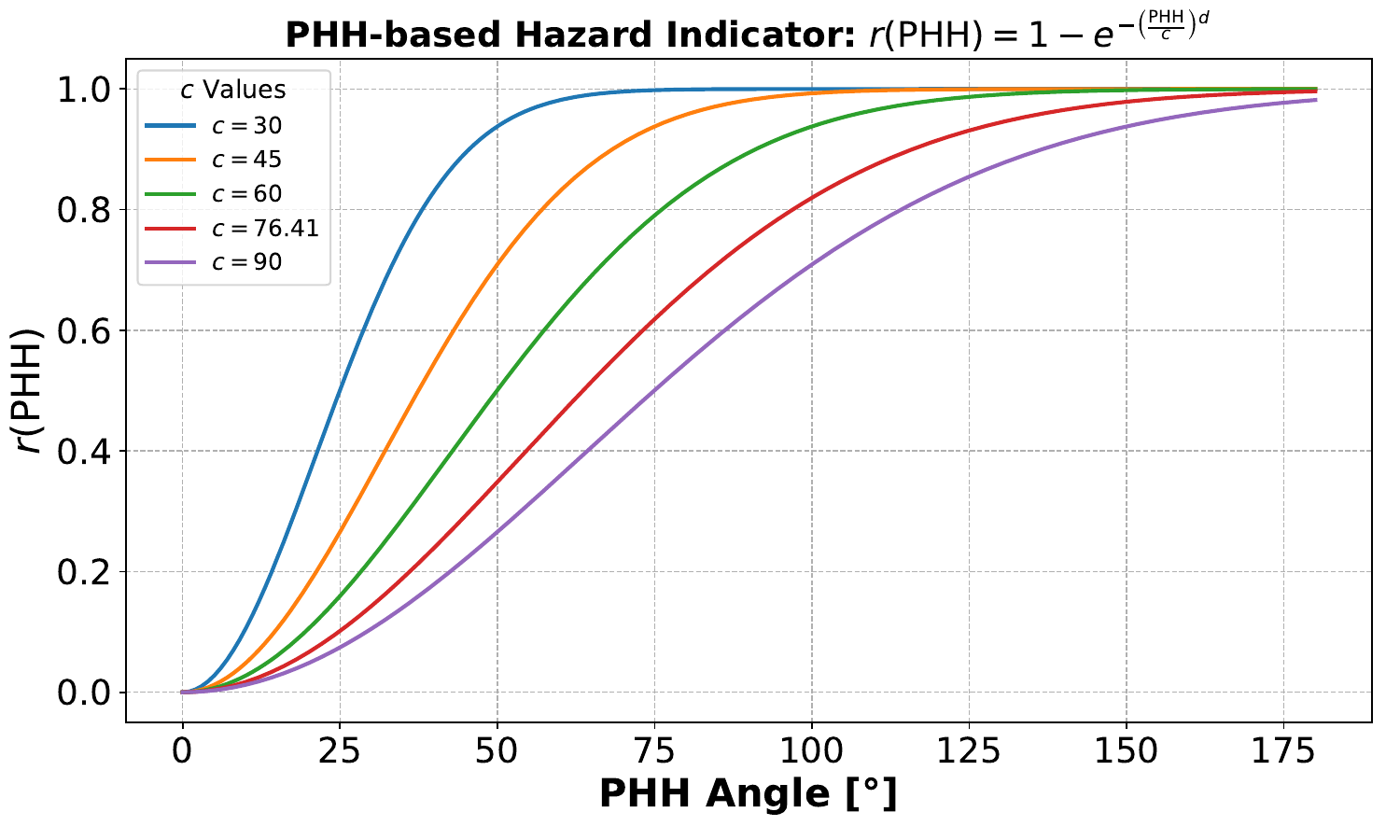}
    \caption{Plot of the PHH-based hazard indicator for different values of $c$.}
    \label{fig:PHH_Metric}
\end{figure}
When the human operator directly faces the robot $PHH = \SI{0}{\degree}$, and the hazard is minimal. As the head rotates away, the hazard increases, maxing out at $PHH = 1$ at an angle of $90\si{\degree}$ meaning the heads orientation is perpendicular to the robot. For angles beyond $90\si{\degree}$, up to $180\si{\degree}$, remains one, indicating the operator cannot observe the robot. Beyond this point, the angle is irrelevant, as the situation is considered unsafe. Ultimately, the function ensures rapid convergence to the maximum value beyond a critical angle.

The hazard indicator is illustrated in Fig. \ref{fig:PHH_Metric} for different values of $c$. Adjusting $c$ is critical to ensure the hazard evaluation aligns with safety requirements and practical expectations, balancing sensitivity and usability.

\subsubsection{Total Hazard Indicator}

The total hazard indicator is computed, by taking the weighted sum of all considered indicators, and dividing it by the sum of the corresponding weights. The corresponding equation is:

\begin{equation}
    R_{\text{total}} = \frac{1}{\sum_{i=1}^{3} \omega_i} \sum_{i=1}^{3} \omega_i \cdot r_i
\end{equation}

with 
$r_1 = r_\mathrm{D}(d_\mathrm{H})$, $r_2 = r_\mathrm{V}(v,\theta)$, $r_3 = r_{\text{PHH}}(PHH)$.
To offset the fact that there is only one anthropocentric parameter, it is weighted with double the value of the other two weights. This results in $\omega_1 = \omega_2 = 1$ and $\omega_3 = 2$. The total hazard indicator has only a non-zero value, if both velocity $v \geq V_\mathrm{min}$ and distance $ d \geq d_\mathrm{reach}$.

\section{Case Study} \label{sec:case_study}

The presented methodologies are evaluated on a dataset, named ABiD, depicting HRC scenarios in industrial context which is set to be published soon. It includes three-dimensional point clouds, two-dimensional images, robot joint states, joint velocities, effort, and the spatial layout of sensors and actors in the scene. Due to the labor-intensive nature of manual labeling, an automated labeling process is used, followed by manual corrections, resulting in the generation of 2D and 3D labels.

\begin{figure}[!htpb]
    \epsfxsize=.001\hsize
    \centerline{\epsfbox{ch01f01.eps}}
    \includegraphics[width=\linewidth]{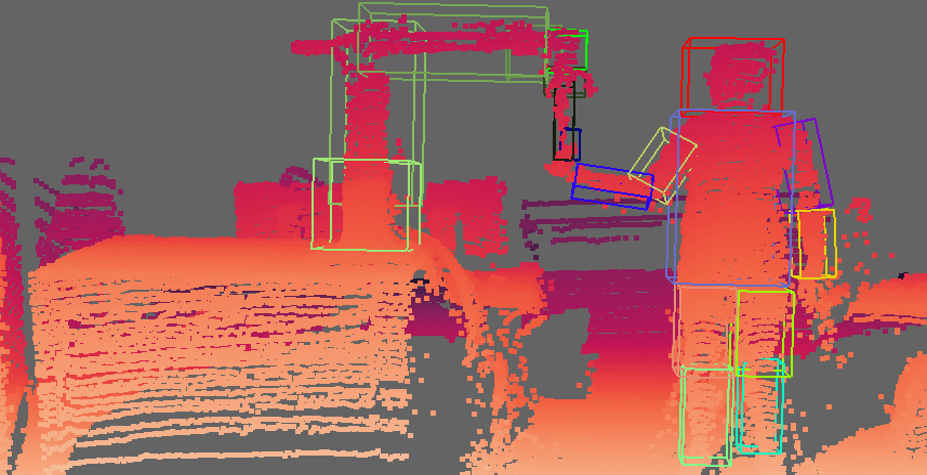}
    \caption{Example taken from the ABiD Dataset showcasing a handover scenario. \textcopyright Proximity Robotics \& Automation GmbH}
    \label{fig:abid_example}
\end{figure}

The dataset comprises three scenarios each recorded in both dangerous and non-dangerous conditions. In the \textbf{Collaboration scenario}, humans and robots collaborate on a shared task. In the \textbf{Handover scenario}, the robot transfers an object to a human. In the \textbf{Coexistence scenario}, humans and robots work in proximity without direct interaction. Fig. \ref{fig:abid_example} depicts an example point cloud of the dataset used for the case study.
The hazard analysis results are presented for dangerous and non-dangerous handover scenarios, both involving a UR10 cobot moving an object toward a human operator for retrieval. Each frame undergoes hazard analysis to quantify the inherent danger of the scenarios. This includes calculating distances between all human limbs and all robot links, determining robot velocity using forward kinematics, and assessing human head orientation from point cloud data. 

\begin{figure}[!htpb]
    \epsfxsize=.001\hsize
    \centerline{\epsfbox{ch01f01.eps}}
    \includegraphics[width=\linewidth]{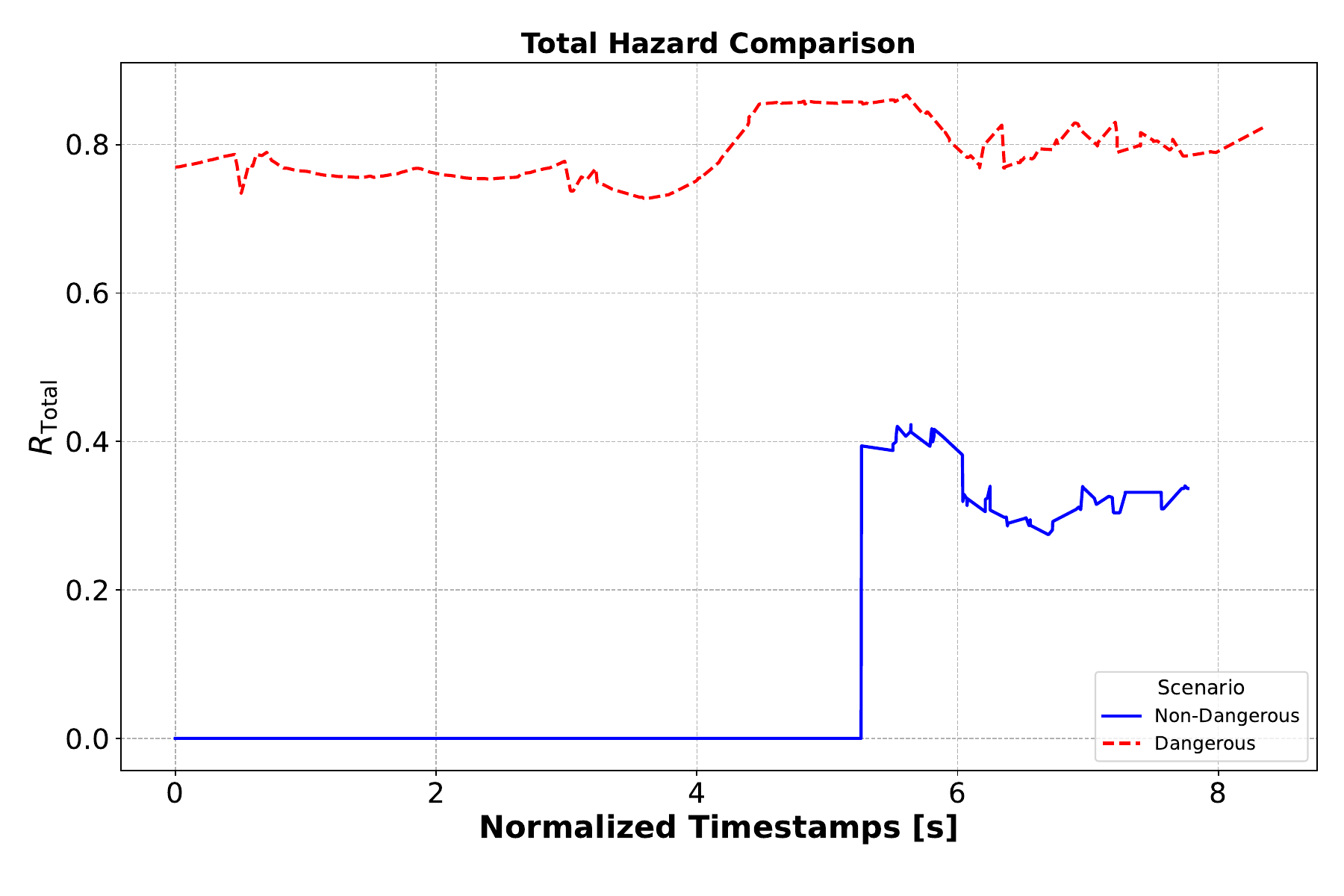}
    \caption{Comparison of the total hazard for the handover scenarios.}
    \label{fig:total_hazard}
\end{figure}

\begin{figure}[!htpb]
    \epsfxsize=.001\hsize
    \centerline{\epsfbox{ch01f01.eps}}
    \includegraphics[width=\linewidth]{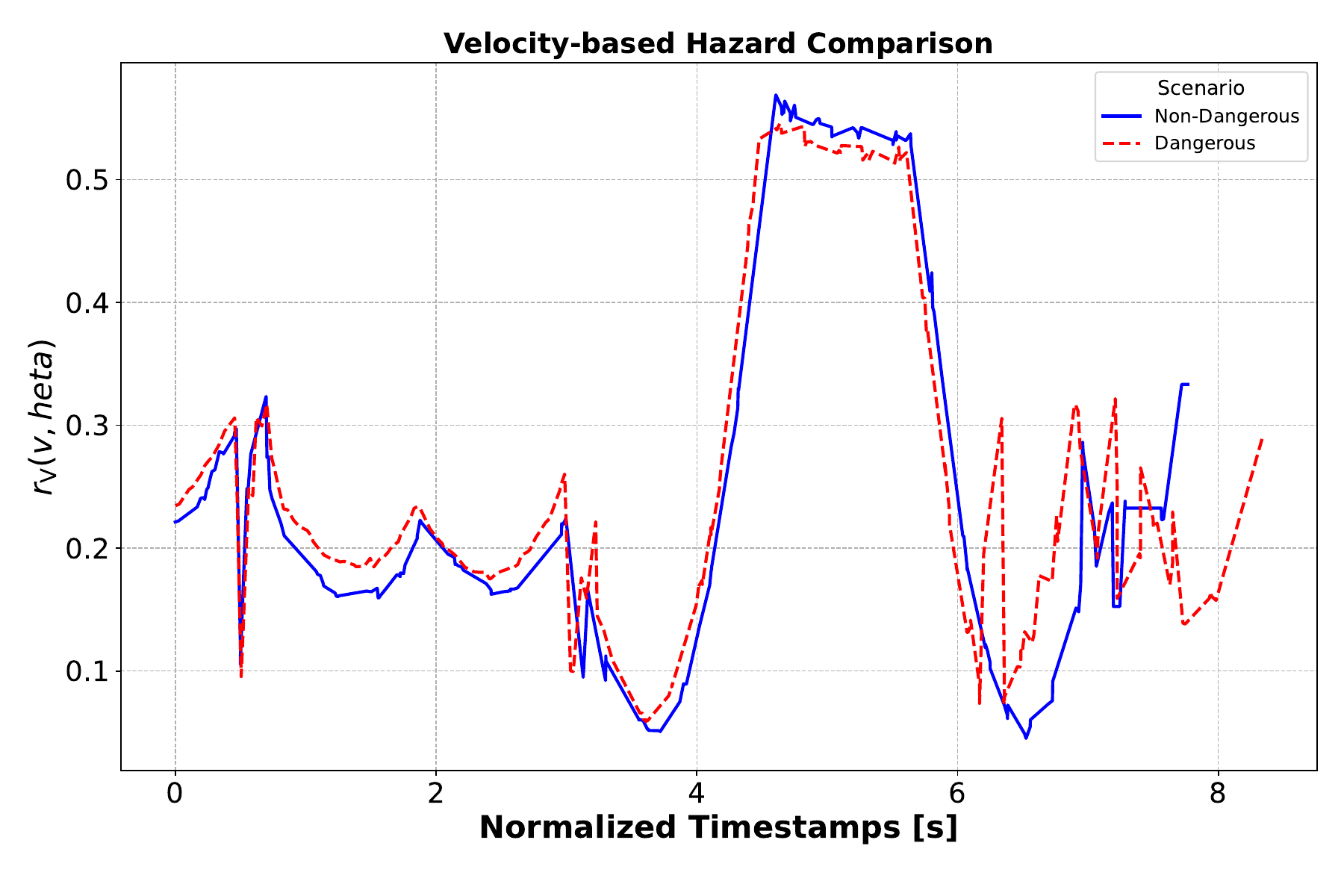}
    \caption{Comparison of velocity-based hazard for the handover scenario.}
    \label{fig:velocity_hazard_comparison}
\end{figure}

First, the total hazard of the two scenarios is compared, as depicted in Fig. (\ref{fig:total_hazard}). It is evident, that the dashed red lines illustrate the dangerous scenario, while the solid blue lines represent the non-dangerous scenario. However, this high-level overview on the hazard does not provide much contextual information. Without prior knowledge of the individual hazard indicators, it cannot be pinpointed where the danger stems from. Further analysis is needed to identify and mitigate factors contributing to this.
Fig. \ref{fig:velocity_hazard_comparison} presents the velocity-based hazards for the two scenarios. The hazard level for both scenarios are almost identical. This is to be expected as the two scenarios model identical robot movement. The minor differences are attributable to minor errors in computations of the velocity. The maximal hazard value is approx. $0.55$ for both scenarios. Overall, the velocity of the robot is not responsible for the hazard.

\begin{figure}[!htpb]
    \epsfxsize=.001\hsize
    \centerline{\epsfbox{ch01f01.eps}}
    \includegraphics[width=\linewidth]{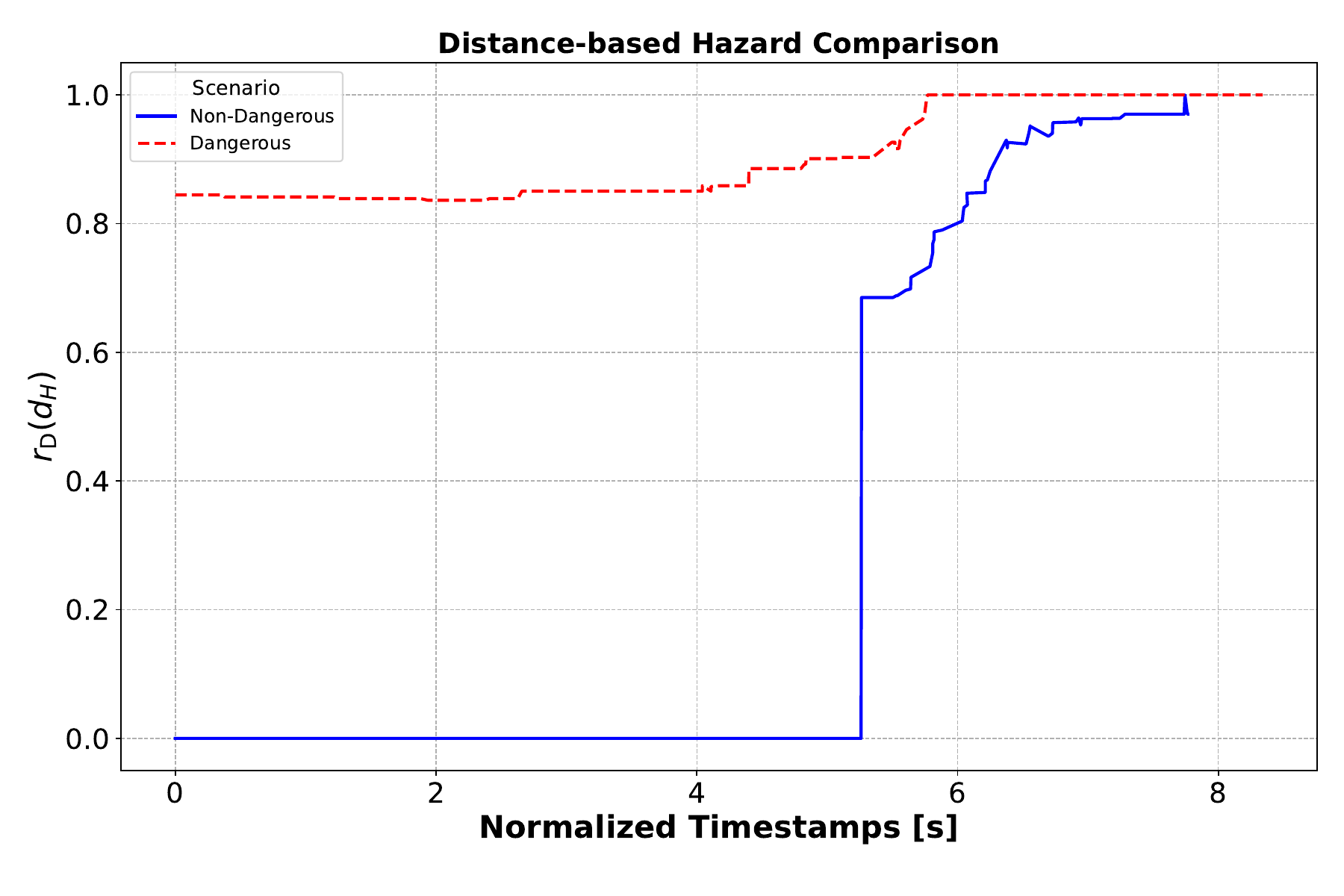}
    \caption{Comparison of distance-based hazard for the handover scenarios.}
    \label{fig:distance hazard comparison}
\end{figure}

Comparing the distance-based hazards, as illustrated in Fig. \ref{fig:distance hazard comparison}. The primary differences occur during the initial few seconds. In the non-dangerous scenario, the human operator remains outside the robot's reach. Conversely, in the dangerous scenario, the operator is positioned much closer to the robot, resulting in a higher hazard indicator. Over time, the metric increases in both scenarios, reaching a maximal value as the minimum safe distance is reached during the handover. However, in the dangerous scenario, this elevated hazard persists for a significantly longer duration. This disparity warrants further investigation during risk assessment to determine why the hazard value remains consistently high in the dangerous scenario, as it may indicate a malfunction, such as the robot failing to move away from the human, or another underlying risk.

\begin{figure}[!htpb]
    \epsfxsize=.001\hsize
    \centerline{\epsfbox{ch01f01.eps}}
    \includegraphics[width=\linewidth]{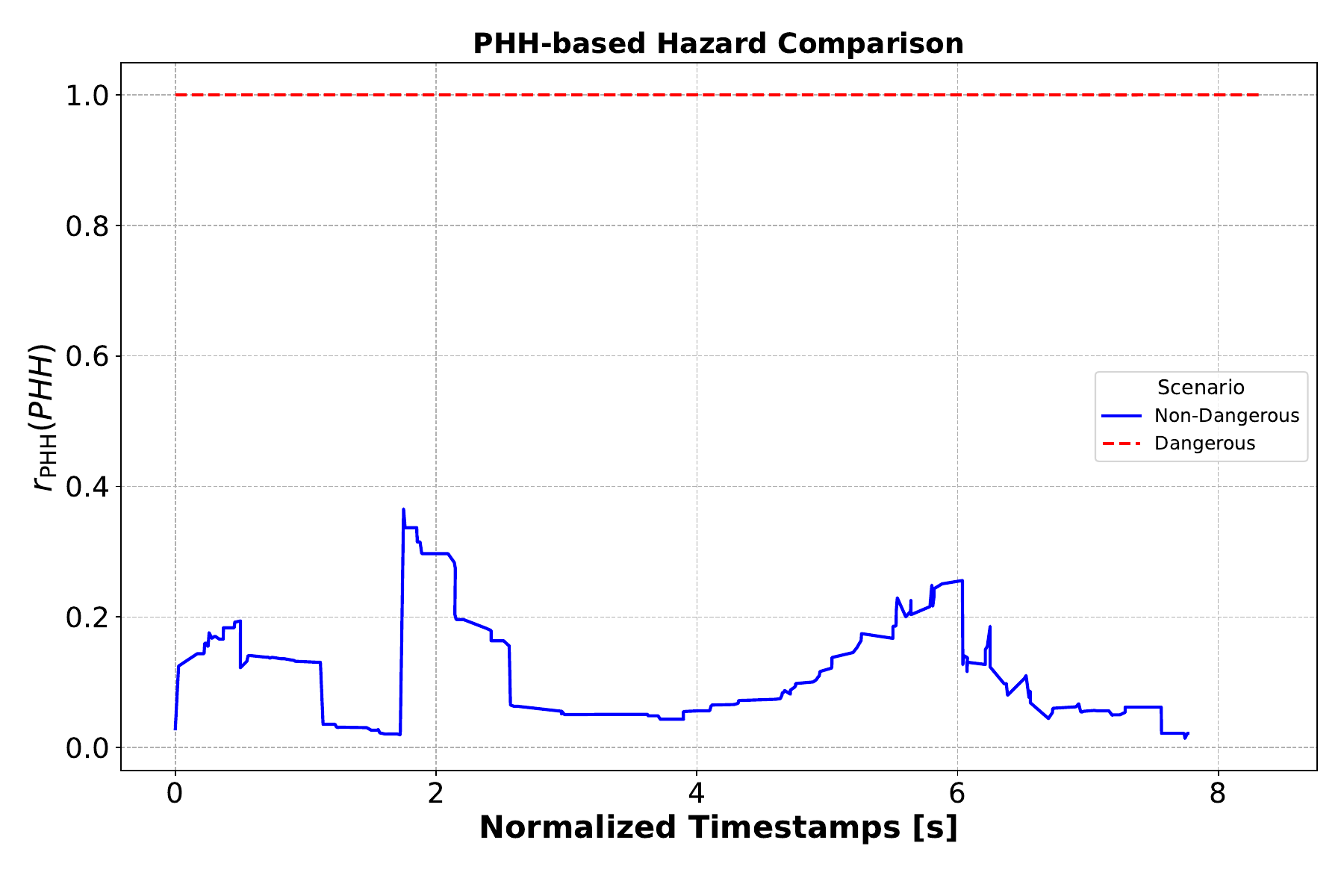}
    \caption{Comparison of PHH-based hazard for the handover scenario.}
    \label{fig:phh-based hazard}
\end{figure}

Finally, the PHH-based comparison demonstrates the most substantial differences, as shown in Fig. \ref{fig:phh-based hazard}. In the dangerous scenario, the indicator is almost constantly maximal, whereas in the non-dangerous scenario, it remains near zero. This contrast arises because, in the non-dangerous scenario, the human operator maintains focus on the task, with only minor deviations in the orientation. In contrast, in the dangerous scenario, the operator looks away and is distracted.

The analysis underscores that even inherently non-dangerous scenarios can become hazardous due to human behavior. Furthermore, it highlights the importance of considering all individual indicators, rather than relying on one value for making decisions, when performing a hazard analysis in the context of dynamic risk assessment.

\section{Conclusion and Future Work} \label{sec:conclusion}

The main contribution of this paper is a hazard analysis method for dynamic risk assessment of HRC scenarios. Scene parameters and an anthropocentric parameter are introduced, and their associated hazards are quantified using heuristic risk functions. This methodology is
applied to analyze hazards in an industrial HRC dataset, revealing that identical scenarios can yield different hazard outcomes depending on human behavior. The analysis emphasizes the importance of a holistic approach that considers all parameters before deciding whether mitigation measures are necessary. Future work can enhance this methodology by incorporating additional scene parameters, such as noise levels, robot joint forces, and torques, to better address collision scenarios and hand-guiding tasks. Additionally, integrating measures of uncertainty could improve the robustness of the analysis.

\bibliographystyle{chicago}
\bibliography{References}

\begin{thebibliography}{}

\bibitem[\protect\citeauthoryear{Aaltonen and Salmi}{Aaltonen and Salmi}{2019}]{aaltonen2019experiences}
Aaltonen, I. and T.~Salmi (2019).
\newblock Experiences and expectations of collaborative robots in industry and academia: Barriers and development needs.
\newblock {\em Procedia Manufacturing\/}~{\em 38}, 1151--1158.

\bibitem[\protect\citeauthoryear{Bdiwi, Pfeifer, and Sterzing}{Bdiwi et~al.}{2017}]{bdiwi2017new}
Bdiwi, M., M.~Pfeifer, and A.~Sterzing (2017).
\newblock A new strategy for ensuring human safety during various levels of interaction with industrial robots.
\newblock {\em CIRP Annals\/}~{\em 66\/}(1), 453--456.

\bibitem[\protect\citeauthoryear{Beltran, Diwa, Gales, Perez, Saguisag, and Serrano}{Beltran et~al.}{2018}]{beltran2018fuzzy}
Beltran, E.~P., A.~A.~S. Diwa, B.~T.~B. Gales, C.~E. Perez, C.~A.~A. Saguisag, and K.~K.~D. Serrano (2018).
\newblock Fuzzy logic-based risk estimation for safe collaborative robots.
\newblock In {\em 2018 IEEE 10th International Conference on Humanoid, Nanotechnology, Information Technology, Communication and Control, Environment and Management (HNICEM)}, pp.\  1--5. IEEE.

\bibitem[\protect\citeauthoryear{Dittrich and Woern}{Dittrich and Woern}{2016}]{dittrich2016robot}
Dittrich, F. and H.~Woern (2016).
\newblock Robot activity adaptation for safe human-robot collaboration based on probabilistic risk modeling.
\newblock In {\em 2016 18th Mediterranean Electrotechnical Conference (MELECON)}, pp.\  1--6. IEEE.

\bibitem[\protect\citeauthoryear{Huck, Ledermann, and Kröger}{Huck et~al.}{2022}]{Huck_Testing_System_Safety}
Huck, T.~P., C.~Ledermann, and T.~Kröger (2022).
\newblock Testing robot system safety by creating hazardous human worker behavior in simulation.
\newblock {\em IEEE Robotics and Automation Letters\/}~{\em 7\/}(2), 770--777.

\bibitem[\protect\citeauthoryear{Huck, Münch, Hornung, Ledermann, and Wurll}{Huck et~al.}{2021}]{Huck_Risk_assesment_survey}
Huck, T.~P., N.~Münch, L.~Hornung, C.~Ledermann, and C.~Wurll (2021).
\newblock Risk assessment tools for industrial human-robot collaboration: Novel approaches and practical needs.
\newblock {\em Safety Science\/}~{\em 141}, 105288.

\bibitem[\protect\citeauthoryear{{ISO 10218-2}}{{ISO 10218-2}}{2011}]{ISO10218}
{ISO 10218-2} (2011).
\newblock Robots and robotic devices - safety requirements for industrial robots - part 2: Robot systems and integration.
\newblock Technical Report ISO 10218-2:2011, International Organization for Standardization.
\newblock [Online]. Available: \url{http://www.iso.org}.

\bibitem[\protect\citeauthoryear{{ISO 15066}}{{ISO 15066}}{2016}]{ISO15066}
{ISO 15066} (2016).
\newblock Robots and robotic devices - collaborative robots.
\newblock Technical Report ISO/TS 15066:2016, International Organization for Standardization.
\newblock [Online]. Available: \url{http://www.iso.org}.

\bibitem[\protect\citeauthoryear{{ISO/TR 14121-2}}{{ISO/TR 14121-2}}{2012}]{ISO14121}
{ISO/TR 14121-2} (2012).
\newblock Safety of machinery — risk assessment — part 2: Practical guidance and examples of methods.
\newblock Technical Report ISO/TR 14121-2:2012, International Organization for Standardization.
\newblock [Online]. Available: \url{http://www.iso.org}.

\bibitem[\protect\citeauthoryear{Kaplan and Garrick}{Kaplan and Garrick}{1981}]{kaplan1981quantitative}
Kaplan, S. and B.~J. Garrick (1981).
\newblock On the quantitative definition of risk.
\newblock {\em Risk analysis\/}~{\em 1\/}(1), 11--27.

\bibitem[\protect\citeauthoryear{Kulić and Croft}{Kulić and Croft}{2006}]{Kulic_real_time_safety_for_HRI}
Kulić, D. and E.~A. Croft (2006).
\newblock Real-time safety for human–robot interaction.
\newblock {\em Robotics and Autonomous Systems\/}~{\em 54\/}(1), 1--12.

\bibitem[\protect\citeauthoryear{Marvel and Norcross}{Marvel and Norcross}{2017}]{marvel2017implementing}
Marvel, J.~A. and R.~Norcross (2017).
\newblock Implementing speed and separation monitoring in collaborative robot workcells.
\newblock {\em Robotics and computer-integrated manufacturing\/}~{\em 44}, 144--155.

\bibitem[\protect\citeauthoryear{{Next Move Strategy Consulting}}{{Next Move Strategy Consulting}}{2022}]{next_move_strategy_2022}
{Next Move Strategy Consulting} (2022).
\newblock Size of the collaborative robot (cobot) end-effector market worldwide in 2020 and 2021, with a forecast for 2022 to 2030 (in million u.s. dollars) [graph].
\newblock In Statista.
\newblock Retrieved January 10, 2025, from \url{https://www.statista.com/statistics/1292113/cobot-end-effector-market-size-worldwide/}.

\bibitem[\protect\citeauthoryear{Sanderud, Thomessen, Osumi, and Niitsuma}{Sanderud et~al.}{2015}]{sanderud2015proactive}
Sanderud, A., T.~Thomessen, H.~Osumi, and M.~Niitsuma (2015).
\newblock A proactive strategy for safe human-robot collaboration based on a simplified risk analysis.
\newblock {\em Modeling, Identification and Control\/}~{\em 36}.

\bibitem[\protect\citeauthoryear{Zacharaki, Kostavelis, Gasteratos, and Dokas}{Zacharaki et~al.}{2020}]{zacharaki2020safety}
Zacharaki, A., I.~Kostavelis, A.~Gasteratos, and I.~Dokas (2020).
\newblock Safety bounds in human robot interaction: A survey.
\newblock {\em Safety science\/}~{\em 127}, 104667.

\end{thebibliography}
\end{document}